%% file: main.tex
\documentclass[conference]{IEEEtran}
\IEEEoverridecommandlockouts

\usepackage[table,xcdraw]{xcolor}
\usepackage{cite}
\usepackage[edges]{forest}
\usepackage{tikz}
\usepackage{amsmath,amssymb,amsfonts}
\usepackage{algorithmic}
\usepackage{graphicx}
\usepackage{textcomp}
\usepackage{adjustbox}

\usepackage[none]{hyphenat}
\usepackage{comment}
\usepackage{array}
\usepackage{caption}
\usepackage{multirow}
\newcolumntype{P}[1]{>{\centering\arraybackslash}p{#1}}
\newcolumntype{M}[1]{>{\centering\arraybackslash}m{#1}}
\def\BibTeX{{\rm B\kern-.05em{\sc i\kern-.025em b}\kern-.08em
    T\kern-.1667em\lower.7ex\hbox{E}\kern-.125emX}}

\begin{document}

\title{
Large Language Models for Financial Aid \\in Financial Time-series Forecasting
}

\author{
    \IEEEauthorblockN{
    Md Khairul Islam \textsuperscript{\rm 1},
    Ayush Karmacharya \textsuperscript{\rm 1},
    Timothy Sue \textsuperscript{\rm 1},
    Judy Fox \textsuperscript{\rm 1,2}
    }

    \IEEEauthorblockA{
        \textsuperscript{\rm 1} Computer Science Department, University of Virginia \\
     \textsuperscript{\rm 2} School of Data Science, University of Virginia
      \\
      Charlottesville, USA \\
    Email : \{mi3se, psb7wm, und2yw, cwk9mp\}@virginia.edu 
    }
}


\maketitle

\begin{abstract}
Considering the difficulty of financial time series forecasting in financial aid, much of the current research focuses on leveraging big data analytics in financial services. One modern approach is to utilize "predictive analysis", analogous to forecasting financial trends. However, many of these time series data in Financial Aid (FA) pose unique challenges due to limited historical datasets and high dimensional financial information, which hinder the development of effective predictive models that balance accuracy with efficient runtime and memory usage. Pre-trained foundation models are employed to address these challenging tasks. We use state-of-the-art time series models including pre-trained LLMs (GPT-2 as the backbone), transformers, and linear models to demonstrate their ability to outperform traditional approaches, even with minimal ("few-shot") or no fine-tuning ("zero-shot"). Our benchmark study, which includes financial aid with seven other time series tasks, shows the potential of using LLMs for scarce financial datasets.
\end{abstract}


\section{Introduction}

The advancement of AI has taken over many domains including the field of financial market \cite{cao2022ai} and big data \cite{zu2023finformer}. In particular, financial time series forecasting has improved significantly from using statistical models to machine learning \cite{leung2021financial} and then deep learning \cite{li2024finreport}. These financial forecasting areas include currency exchange rate \cite{sako2022neural, wu2023timesnet}, stock market \cite{zu2023finformer}\cite{sako2022neural}\cite{li2024finreport}, commodity prices \cite{xu2022commodity}\cite{zhang2023oil} and more. \textbf{Pre-trained foundation models}, such as large language models (LLMs) have driven the progress in Natural Language Processing (NLP) and Computer Vision (CV). Foundation models like GPT \cite{radford2019language}, and Vision Transformer \cite{dosovitskiy2020image} can perform well on a diverse range of tasks in few-shot (little training) or zero-shot (no training) learning. This enables applications where historical data is limited or mostly missing.

\textbf{Financial time series forecasting (FTSF)} is an important domain that needs more attention among the multivariate time series forecasting tasks. Previous works on FTSF rely heavily on machine learning \cite{leung2021financial} or traditional deep learning \cite{li2024finreport} methods. With some recent works on multi-modal FTSF \cite{chen2023chatgpt}\cite{yu2023temporal}. \textbf{Financial aid (FA)} is crucial to many students' educational journeys, providing the resources needed to pursue academic dreams while fostering educational equity. However, the process, including tasks like processing the free application for Federal Student Aid (FAFSA), is often manual, time-consuming, and susceptible to errors. Access to historical datasets is limited to yearly intervals and is subject to changes in policy.
\input{tables/dataset}
 We describe and evaluate LLM-based foundation models in the FTSF domain using 8 deep-learning models and compare especially the financial aid with 7 other financial datasets. Our research questions are,
\begin{itemize}
    \item\textit{Q1}: Which models are better as few-shot learners?
    \item\textit{Q2}: Can pre-trained LLMs perform zero-shot learning?
\end{itemize}

Answering these questions will help us better understand the current advancement of LLMs for financial time series forecasting. In summary, our contributions are, 
\begin{itemize}
    \item Collect eight datasets from four financial domains (Stock, Commodity, Currency, Institution) for over 10 years.
    \item Benchmark five state-of-the-art time series deep learning models, and three LLM-based foundation models on these datasets. 
    \item Open source code and datasets at GitHub \footnote{https://github.com/UVA-MLSys/Financial-Time-Series} to facilitate full reproducibility and further research in this domain.
\end{itemize}


\section{Methodology \label{sec:methodology}}

\subsection{Problem Statement}
Given the input dataset, $X \in \mathcal{R}^{F \times T} $, $T$ denotes the total timesteps in days and $F$ input features (including past targets and other features). With a lookback window of $L$ past days, the input at time $t$ is $X_t = X_{t-(L-1):t}$ which contains inputs of the last $L$ days. Given this input $X_t$, the model $f$ predicts the targets $O$ (e.g. stock prices) for the next $\tau_{max}$ days. The target output $y_t$ at time $t$ can be expressed as,

\begin{equation}
\begin{aligned}
   \hat{y}_{t} & = f(X_t), \text{where, }\\
  X_t &= x_{t-(L-1):t} = [x_{t-(L-1)}, x_{t-(L-2)}, \cdots, x_t]\\
&= \{ x_{f, l, t}\}, ~ f \in \{1, \cdots, F\}, ~ l \in \{1, \cdots, L\}
\end{aligned}
\end{equation}

For the financial aid data, funds allocated to each state are a time series with $T$ years (2004 to 2020). A lookback $ L $ of 10 years is used to predict funds ($O$) for the next year ($\tau_{max} = 1$). For all other datasets, we have daily inputs for 10 years. With a lookback window of the past 96 days ($L = 96$), we predict the targets for the next 24 days ($\tau_{max}=24$).

\subsection{Dataset}
\begin{figure}[!htbp]
\centering
\includegraphics[width=0.35\textwidth]{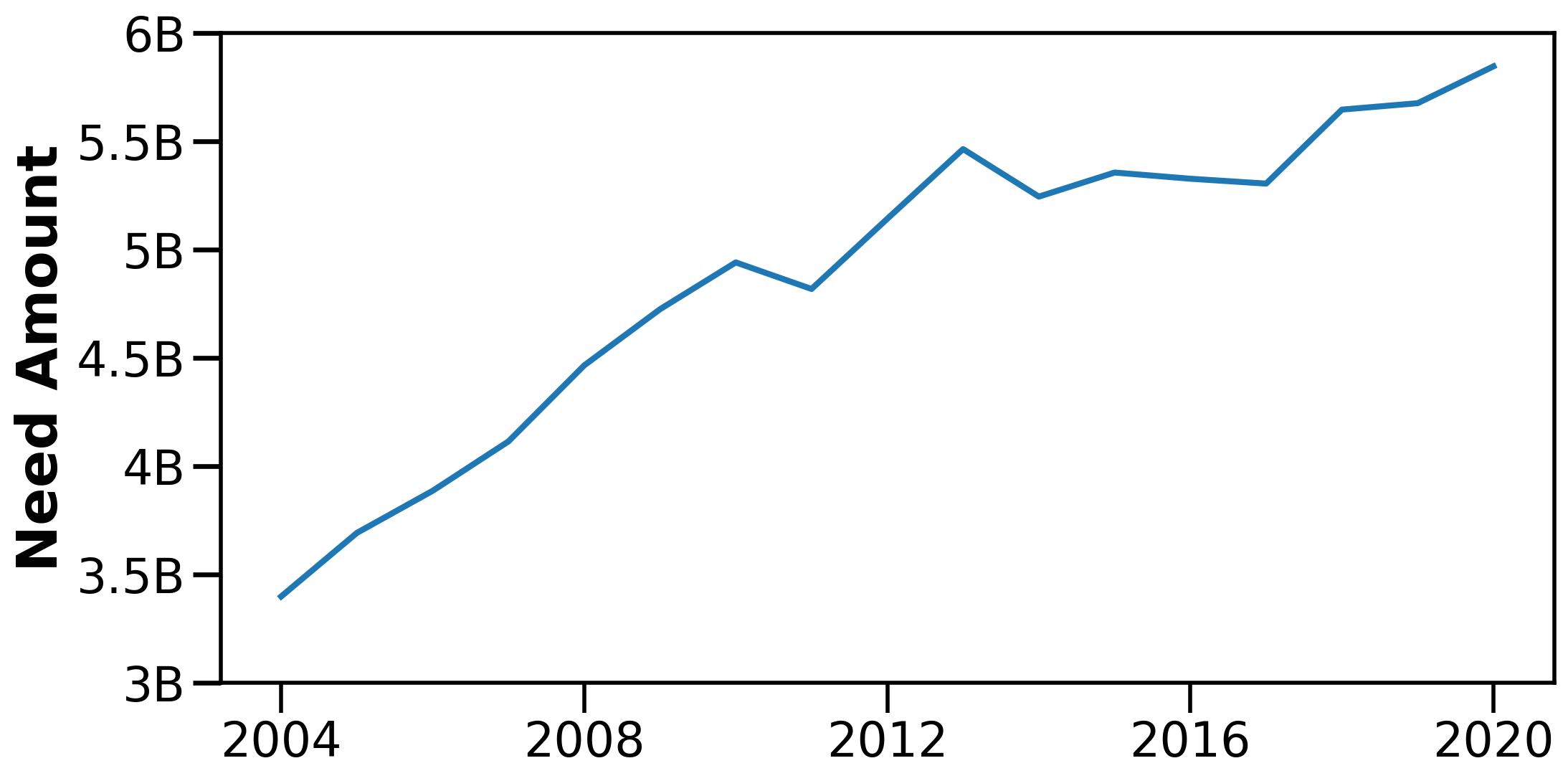}
\caption{Financial Aid data aggregated at the state level from 2004 to 2020 (17 years), in billions of US dollars. Access to historical datasets is limited to yearly intervals.}
\label{fig:financial_aid}
\end{figure}

We use the following financial datasets:  (1) \textbf{Financial Aid}: Financial aid distributed to each US state by the Government to support student education and collected from years 2004 to 2020 \cite{financial_aid}. Details of available features are in Table \ref{tab:financial_aid} and the yearly aggregated aid in Fig \ref{fig:financial_aid}. (2) \textbf{Stock Market} \cite{sako2022neural}\cite{yu2023temporal}\cite{li2024finreport}\cite{gruver2024large}: Includes the daily stock prices (Close, Open, High, Low) and volumes for each of the following stocks up to 10 years from the NASDAQ database: S\&P 500 (SPX), Microsoft Corporation (MSFT), and Apple Inc (AAPL); (3) \textbf{Commodities} \cite{xu2022commodity}\cite{zhang2023oil}: Contains data (Close, Open, Volume, High, Low) on different kinds of raw materials such as Natural Gas, Crude Oil and Gold.
(4) \textbf{Currency Exchange Rate} \cite{sako2022neural}\cite{wu2023timesnet}: The currency units per U.S. dollar reported daily by the issuing central bank (rates are not recorded on the weekends and certain holidays). This data covers the following currencies: Australian Dollar (AUD), Canadian Dollar (CAD), Chinese yuan (CNY), Euro (EUR), Indian rupee (INR), Japanese yen (JPY), and the U.K. pound (GBP); 

The statistics are in Table \ref{table:dataset}. The train, validation, and test split follows the 8:1:1 ratio, the validation set follows the train set, then the test set. The data is standard normalized before passing to the model. The few missing values ($<$1\%) are imputed using last-seen valid values.

\begin{table}[htbp]
    \centering
    \caption{List of available features in financial aid \cite{financial_aid}. Aid is given based on financial needs, academic merit, or both. The sub-categories are simplified and describe multiple features.}
    \begin{tabular}{|p{1cm}|p{1.5cm}|p{5cm}|} \hline
    \textbf{Category} & \textbf{Sub-category} & \textbf{Description} \\\hline
        \multirow{10}{=}{Need, Merit, both } & Identifier & State id and name abbreviation.\\ \cline{2-3}
        & Number & Total students receiving the award.\\ \cline{2-3}
        & Public/Private & Whether the funds can be used for public or private sectors and how long (2 or 4 years).\\\cline{2-3}
        & Flags & 0 or 1 based on whether the aid falls in a particular category. \\\cline{2-3}
        & Program & Aid program with the most generous eligibility criteria.\\\cline{2-3}
        & Notes & Related text. \\\cline{2-3}
        & Threshold & GPA, SAT, income, and other academic or financial limits to qualify for the aid. \\ \hline
     
        Time & Year & Fiscal or academic year. \\ \hline
        Target & Amount & Aid amount received by the students.\\
         \hline
    \end{tabular}
    
    \label{tab:financial_aid}
\end{table}

\subsection{Models}

\begin{figure}[!htp]
\centering
\includegraphics[width=0.47\textwidth, height=0.37\textwidth]{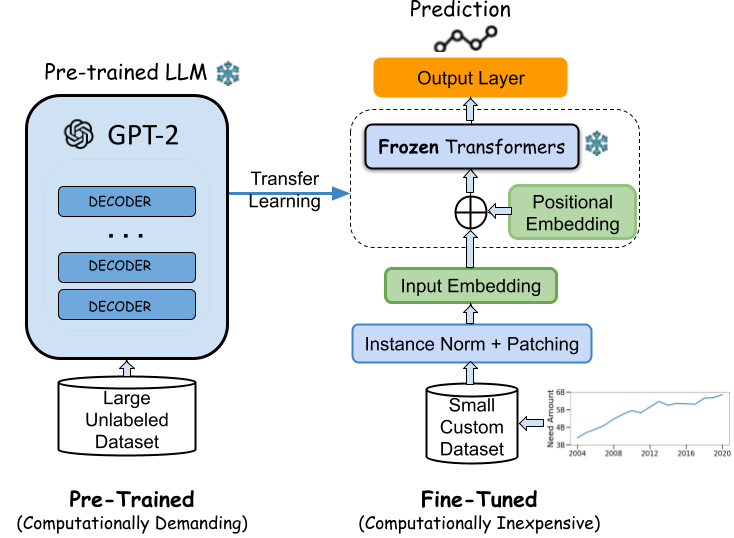}
\caption{A high-level overview of pre-training an LLM and fine-tuning on a custom dataset (e.g. the Financial Aid dataset) for downstream tasks.}
\label{fig:pretrainedllm}
\end{figure}

We use the following time series models in our work. The models are chosen based on their popularity and recently published work. We focus on point forecasting in our work. A high-level overview of how pre-trained LLMs are fine-tuned for custom datasets is illustrated in Figure \ref{fig:pretrainedllm}.


\subsubsection{\textbf{LLM Foundation Models}} 

The LLM-based foundation models are selected based on their versatility in time series forecasting.  We use the configurations from \cite{tan2024language}. The pre-trained foundation models are frozen except for the last layer when fine-tuning. These models use a pre-trained GPT-2 \cite{radford2019language} as the LLM backbone. We select the following recent models: (1) TimeLLM \cite{jin2023time} (2) CALF \cite{liu2024taming} (3) GPT4TS (One Fits All, \cite{zhou2023one}).


\subsubsection{\textbf{Traditional Models}}
We choose the following recent non-pre-trained models: (1) DLinear \cite{Zeng2022AreTE} (2) iTransformer \cite{liu2024itransformer} (3) TimesNet \cite{wu2023timesnet} (4) PatchTST \cite{nie2022time} (5) TimeMixer \cite{wang2024timemixer}. Most of these models are Transformer-based and have shown great performance in capturing temporal patterns.

\input{tables/rq2}

\subsection{Implementation Details}

We use the PyTorch framework and follow \cite{wu2023timesnet}\cite{tan2024language} to implement our experiments. Each experiment runs three times with different random seeds (648, 506, 608), and the average results are presented. Following \cite{zhou2023one}\cite{liu2024taming} we use the \textbf{pre-trained GPT2} as the backbone for the LLMs and only fine-tune the output layer during training shown in Fig. \ref{fig:pretrainedllm}. The traditional models are trained from scratch. We use the Adam optimizer with a learning rate 1e-3 and a dropout rate 0.1. The GPT4TS \cite{zhou2023one} model uses the L1 loss. The CALF \cite{liu2024taming} model uses the weighted average of task, feature, and logit loss using the L1 loss. The other models use the Mean Squared Error loss. Models are trained for 10 epochs max, with batch size 32. The experiments were run on an NVIDIA 2080Ti GPU with 11GB+ memory with 32GB RAM. Following \cite{wang2024timemixer}\cite{liu2024taming}\cite{gruver2024large}, we use Mean Square Error (MSE) and Mean Absolute Error (MAE) as evaluation metrics. \textbf{Lower is better} for these metrics. 




\section{Experiments and Results \label{sec:experiments}}

In this section, we investigate the research questions, the setup, and the results.  





\subsection{\textbf{Q1. Which models are better as few-shot learners?}}
Pre-trained models are preferred largely due to their generalizability and good performance in few-shot or zero-shot learning settings \cite{liang2024foundation}. Since these LLMs are already trained on many datasets, they often outperform the other models when few training data are available \cite{liu2024taming} or without training \cite{gruver2024large}. Following \cite{liu2024taming}, we select the last 10\% training data to train the models in a few-shot learning setting. 

\textbf{Results.} The few-shot learning results are shown in Table \ref{table:rq2_results}. All model performance drops significantly after reducing the train data size. This is due to the models' inability to learn enough temporal patterns from the limited input.  TimeLLM performs the best overall (three best and four 2nd best cases). While PatchTST performs the 2nd best with a close margin (three best and two 2nd best cases). DLinears performance drops significantly as it is a simple linear model. However, overall \textit{the LLMs performed better in the few-shot learning}.

\subsection{\textbf{Q2. Can LLMs perform zero-shot learning in FTSF?}}
Real-world scenarios can often have no available past observations (i.e. new company stock in the market, newly launched product). Having zero-shot learning ability is crucial in forecasting those cases since traditional deep learning time series models are unable to train and forecast those cases. We investigate whether LLMs can effectively assist in those cases. We load the pre-trained LLMs and evaluate them on the test set without fine-tuning. Since no training is done in this part, we exclude the traditional time series models from this analysis. 

\input{tables/rq3}

\textbf{Results.} Table \ref{table:rq3_results} shows the zero-shot results of the LLMs. Compared to \textit{Q1}, the results achieved here are significantly worse. Since each financial data may have distinct temporal patterns, without fine-tuning the LLMs fail to forecast them effectively. We conclude, \textit{LLMs are yet not quite effective for zero-shot learning for financial time series.}

\section{Related Works \label{sec:related_works}}

Deep learning for time series has significantly outperformed machine learning approaches \cite{wang2024timemixer}\cite{liu2024itransformer}, also in finance \cite{li2024finreport}. \cite{sako2022neural} used RNN models to forecast stock market prices, and currency exchange rates. Many recent deep learning models have been used to forecast the stock market \cite{zu2023finformer}\cite{sako2022neural}\cite{li2024finreport}, commodity prices \cite{xu2022commodity}\cite{zhang2023oil}. Foundation models in time series have recently gained significant attention \cite{liang2024foundation}. Pre-trained LLMs and vision models have been enhanced for time series. \cite{gruver2024large}\cite{das2023decoder} showed the ability of LLMs to perform in zero-shot and few-shot settings in time series tasks. GPT4TS \cite{zhou2023one} leverages pre-trained language models without altering important layers. TimeLLM \cite{jin2023time} reprograms LLM's ability to reason with time series data by proposing a prompt-as-prefix technique. CALF \cite{liu2024taming} proposed a novel fine-tuning framework to reduce the distribution discrepancy between textual and temporal data. Chronos \cite{ansari2024chronos} performed significantly in probabilistic forecasting. 


\section{Conclusion and Future Works \label{sec:conclusion}}
In this paper, we benchmark financial datasets from multiple domains using state-of-the-art time series models and LLM-based foundation models. Our results show that LLMs are more effective for few-shot and zero-shot learning. \textbf{Especially, the few-shot and zero-shot capabilities of LLMs can be effective for financial Aid practitioners who are currently unable to apply deep learning methods due to limited data availability}.  We focus on point forecasting with a single modality in this work. Incorporating data from different financial modalities into time series models will be future work, and probabilistic forecasting can help the financial domain by outputting a probabilistic distribution. Our research highlights the potential of foundation LLMs in financial aid and the overall finance time series forecasting domain.

\section*{Acknowledgment}
This work is partly supported by NSF grant CCF-1918626  Expeditions: Collaborative  Research: Global Pervasive Computational Epidemiology, and NSF Grant 2200409 
for CyberTraining:CIC: CyberTraining for Students and Technologies from Generation Z. 

\bibliographystyle{IEEEtran}
\bibliography{bibliography}

\end{document}

%% file: tables/dataset.tex
    

\begin{table*}[!htbp]
    \centering
    \caption{Datasets overview. Time series indicates the number of target time series (i.e., channels). Input features are past observations and dataset size is depicted as training, validation, and test.} 
    \resizebox{\textwidth}{!}{
    \begin{tabular}{|l|l|c|l|c|c|c|c|} \hline
         \textbf{Domain} & \textbf{Dataset} & \textbf{Date} & \textbf{Frequency} & \textbf{Time series} & \textbf{Lookback} & \textbf{Horizon} & \textbf{Size} \\ \hline
          Institution & \cellcolor[HTML]{C3C3C3}Financial Aid & \cellcolor[HTML]{C3C3C3}2004 - 2020 &  \cellcolor[HTML]{C3C3C3}Yearly & \cellcolor[HTML]{C3C3C3}56  & \cellcolor[HTML]{C3C3C3}10 & \cellcolor[HTML]{C3C3C3}1 & \cellcolor[HTML]{C3C3C3}(137, 92, 92) \\
        \hline
         \multirow{3}{*}{Stock} & S\&P 500 & Sep 1, 2014 -  Aug 29, 2024 & Daily & 4 & 96 & 24 & (1903, 231, 229) \\
         & Apple & Sep 2, 2014 -  Aug 29, 2024 & Daily& 5  & 96 & 24 &(1893, 230, 228) \\
          & Microsoft & Sep 3, 2014 -  Aug 30, 2024 & Daily& 4  &96 & 24 &(1893, 230, 228) \\ \hline
          \multirow{3}{*}{Commodity} & Crude Oil & Sep 1, 2014 -  Aug 29, 2024 & Daily & 4 & 96 & 24 &(1893, 230, 228) \\
          & Gold & Sep 1, 2014 -  Aug 29, 2024 & Daily & 4  & 96 & 24 & (1893, 230, 228) \\
           & Natural Gas & Sep 1, 2014 -  Aug 29, 2024 & Daily & 4 & 96 & 24 & (1893, 230, 228) \\ \hline
        Currency & Exchange Rate & Aug 1, 2014 -  Aug 1, 2024 & Daily & 7  & 96 & 24 & (1861, 225, 224) \\ \hline
       
    \end{tabular}
    }
    
    \label{table:dataset}
\end{table*}

%% file: tables/rq2.tex
\begin{table*}[htbp]
    \centering
    \caption{\textit{Q1.} Few-shot learning performance with 10\% training data. TimeLLM and PatchTST 
    outperform the other models. The best and the second best results are in \textcolor[HTML]{000000}
    {\textbf{bold}} and \textcolor[HTML]{000000}{\underline{underlined}}.}
    \begin{adjustbox}{max width=.98\linewidth}
    \begin{tabular}{|p{1.6cm}|cc|cc|cc|cc|cc|cc|cc|cc|} \hline
         \textbf{Method} & \multicolumn{2}{c|}{\textbf{DLinear} \cite{Zeng2022AreTE}} & \multicolumn{2}{c|}{\textbf{PatchTST} \cite{nie2022time}} & \multicolumn{2}{c|}{\textbf{TimesNet} \cite{wu2023timesnet}} & \multicolumn{2}{c|}{\textbf{TimeMixer} \cite{wang2024timemixer}} & \multicolumn{2}{c|}{\textbf{iTransformer} \cite{liu2024itransformer}} &\multicolumn{2}{c|}{\textbf{TimeLLM} \cite{jin2024time}}  & \multicolumn{2}{c|}{\textbf{CALF} \cite{liu2024taming}} & \multicolumn{2}{c|}{\textbf{GPT4TS} \cite{zhou2023one}}  \\ \hline
         \textbf{Metric} & \textbf{MSE} & \textbf{MAE} & \textbf{MSE} & \textbf{MAE} & \textbf{MSE} & \textbf{MAE} & \textbf{MSE} & \textbf{MAE} & \textbf{MSE} & \textbf{MAE} & \textbf{MSE} & \textbf{MAE} & \textbf{MSE} & \textbf{MAE} & \textbf{MSE} & \textbf{MAE}  \\ \hline
Financial Aid  & \cellcolor[HTML]{C3C3C3} 2.94  & \cellcolor[HTML]{C3C3C3} 1.36 & \cellcolor[HTML]{C3C3C3} 2.23  & \cellcolor[HTML]{C3C3C3}\underline{0.95}  & \cellcolor[HTML]{C3C3C3} 2.17  & \cellcolor[HTML]{C3C3C3} 1.04 & \cellcolor[HTML]{C3C3C3} 3.04  & \cellcolor[HTML]{C3C3C3} 1.36 & \cellcolor[HTML]{C3C3C3} 2.22  & \cellcolor[HTML]{C3C3C3} 1.14 & \cellcolor[HTML]{C3C3C3}\underline{2.06}  & \cellcolor[HTML]{C3C3C3}1.08 & \cellcolor{gray!95}\textbf{1.64 }  & \cellcolor{gray!95}\textbf{0.89}  & \cellcolor[HTML]{C3C3C3}2.36 & \cellcolor[HTML]{C3C3C3}1.20 \\  
S\&P 500  & 2.08  & 1.14 & \underline{2.03}  & \textbf{1.14}  & 2.43  & 1.18 & 2.49  & 1.25 & 2.19  & 1.19 & \textbf{1.94 }  & \underline{1.14}  & 2.37 & 1.19 & 3.07 & 1.40 \\ 

Apple  & 2.78  & 1.30 & \underline{2.36}  & \textbf{1.21}  & 3.22  & 1.41 & 3.44  & 1.46 & 3.05  & 1.39 & 3.00 & 1.36 & \textbf{2.33 }  & \underline{1.21}  & 2.80 & 1.29 \\ 

Microsoft  & 2.51  & 1.10 & \textbf{2.13 }  & \textbf{1.06}  & 3.55  & 1.43 & 2.85  & 1.18 & 2.49  & 1.15 & \underline{2.40}  & \underline{1.10}  & 2.97 & 1.27 & 3.40 & 1.37 \\ 

Crude Oil  & \textbf{1.66 }  & \underline{1.01}  & 1.96  & 1.05 & 2.75  & 1.29 & 2.06  & 1.14 & 1.92  & 1.07 & \underline{1.75}  & \textbf{1.01}  & 2.13 & 1.14 & 2.42 & 1.15 \\ 

Gold  & 2.78  & \underline{1.14}  & 2.68  & 1.15 & \underline{2.66}  & 1.17 & \textbf{2.57 }  & \textbf{1.11}  & 3.00  & 1.22 & 2.71 & 1.17 & 3.06 & 1.24 & 3.38 & 1.32 \\ 

Natural Gas  & 2.20  & 1.16 & 2.48  & 1.24 & 2.55  & 1.23 & 2.91  & 1.34 & \textbf{2.12 }  & \textbf{1.12}  & 2.36 & 1.20 & 2.30 & 1.17 & \underline{2.17}  & \underline{1.13}  \\ 

Exchange  & 1.48  & 0.92 & 1.28  & 0.85 & 2.87  & 1.34 & 1.29  & 0.84 & 1.65  & 0.96 & \textbf{1.20 }  & \textbf{0.81}  & 1.48 & 0.92 & \underline{1.22}  & \underline{0.81}  \\ \hline
    \end{tabular}
    \end{adjustbox}
    \label{table:rq2_results}
\end{table*}

%% file: tables/rq3.tex
\begin{table}[!htbp]
    \centering
    \caption{\textit{Q2.} Zero shot performance. GPT4TS performs the best. The best and the second best results are in \textcolor[HTML]{000000}{\textbf{bold}} and \textcolor[HTML]{000000}{\underline{underlined}}. The traditional models are excluded here since they are not pre-trained.}
    \begin{adjustbox}{max width=.45\textwidth}
    \begin{tabular}{|p{1.6cm}|cc|cc|cc|} \hline
         \textbf{Method} &\multicolumn{2}{c|}{\textbf{TimeLLM} \cite{jin2024time}}  & \multicolumn{2}{c|}{\textbf{CALF} \cite{liu2024taming}} & \multicolumn{2}{c|}{\textbf{GPT4TS} \cite{zhou2023one}}  \\ \hline
         \textbf{Metric} & \textbf{MSE} & \textbf{MAE} & \textbf{MSE} & \textbf{MAE} & \textbf{MSE} & \textbf{MAE}  \\ \hline
Financial Aid  & \cellcolor{gray!95}\textbf{2.99 }  & \cellcolor[HTML]{C3C3C3}\textbf{1.29}  & \cellcolor[HTML]{C3C3C3} 3.82  & \cellcolor[HTML]{C3C3C3} 1.59 & \cellcolor[HTML]{C3C3C3}\underline{3.17}  & \cellcolor[HTML]{C3C3C3}\underline{1.41}  \\ 

S\&P 500  & 5.04  & 1.91 & \underline{3.98}  & \textbf{1.74}  & \textbf{3.89 }  & \underline{1.76}  \\ 

Apple  & 4.17  & 1.61 & \underline{3.36}  & \underline{1.44}  & \textbf{3.05 }  & \textbf{1.36}  \\ 

Microsoft  & 5.13  & 1.81 &\underline{4.12}  & \underline{1.62}  & \textbf{3.96 }  & \textbf{1.59}  \\ 

Crude Oil  & 3.05  & 1.39 & \underline{2.21}  & \underline{1.18}  &\textbf{1.89 }  & \textbf{1.08}  \\ 

Gold  & 6.15  & 1.95 & \underline{5.12}  & \textbf{1.76}  & \textbf{5.00 }  & \underline{1.77}  \\ 

Natural Gas  & 4.09  & 1.61 & \underline{3.27}  & \underline{1.43}  & \textbf{2.96 }  & \textbf{1.35}  \\ 

Exchange  & 3.25  & 1.47 & \underline{2.41}  & \underline{1.29}  & \textbf{2.10 }  & \textbf{1.23}  \\ 

\hline

    \end{tabular}
    \end{adjustbox}
    \label{table:rq3_results}
\end{table}

%% file: main.bbl
\begin{thebibliography}{10}
\providecommand{\url}[1]{#1}
\csname url@samestyle\endcsname
\providecommand{\newblock}{\relax}
\providecommand{\bibinfo}[2]{#2}
\providecommand{\BIBentrySTDinterwordspacing}{\spaceskip=0pt\relax}
\providecommand{\BIBentryALTinterwordstretchfactor}{4}
\providecommand{\BIBentryALTinterwordspacing}{\spaceskip=\fontdimen2\font plus
\BIBentryALTinterwordstretchfactor\fontdimen3\font minus \fontdimen4\font\relax}
\providecommand{\BIBforeignlanguage}[2]{{%
\expandafter\ifx\csname l@#1\endcsname\relax
\typeout{** WARNING: IEEEtran.bst: No hyphenation pattern has been}%
\typeout{** loaded for the language `#1'. Using the pattern for}%
\typeout{** the default language instead.}%
\else
\language=\csname l@#1\endcsname
\fi
#2}}
\providecommand{\BIBdecl}{\relax}
\BIBdecl

\bibitem{cao2022ai}
L.~Cao, ``Ai in finance: challenges, techniques, and opportunities,'' \emph{ACM Computing Surveys (CSUR)}, vol.~55, no.~3, pp. 1--38, 2022.

\bibitem{zu2023finformer}
Y.~Zu, J.~Mi, L.~Song, S.~Lu, and J.~He, ``Finformer: A static-dynamic spatiotemporal framework for stock trend prediction,'' in \emph{2023 IEEE International Conference on Big Data (BigData)}.\hskip 1em plus 0.5em minus 0.4em\relax IEEE, 2023, pp. 1460--1469.

\bibitem{leung2021financial}
T.~Leung and T.~Zhao, ``Financial time series analysis and forecasting with hilbert--huang transform feature generation and machine learning,'' \emph{Applied Stochastic Models in Business and Industry}, vol.~37, no.~6, pp. 993--1016, 2021.

\bibitem{li2024finreport}
X.~Li, X.~Shen, Y.~Zeng, X.~Xing, and J.~Xu, ``Finreport: Explainable stock earnings forecasting via news factor analyzing model,'' in \emph{Companion Proceedings of the ACM on Web Conference 2024}, 2024, pp. 319--327.

\bibitem{sako2022neural}
K.~Sako, B.~N. Mpinda, and P.~C. Rodrigues, ``Neural networks for financial time series forecasting,'' \emph{Entropy}, vol.~24, no.~5, p. 657, 2022.

\bibitem{wu2023timesnet}
H.~Wu, T.~Hu, Y.~Liu, H.~Zhou, J.~Wang, and M.~Long, ``Timesnet: Temporal 2d-variation modeling for general time series analysis,'' in \emph{International Conference on Learning Representations}, 2023.

\bibitem{xu2022commodity}
X.~Xu and Y.~Zhang, ``Commodity price forecasting via neural networks for coffee, corn, cotton, oats, soybeans, soybean oil, sugar, and wheat,'' \emph{Intelligent Systems in Accounting, Finance and Management}, vol.~29, no.~3, pp. 169--181, 2022.

\bibitem{zhang2023oil}
S.~Zhang, J.~Luo, S.~Wang, and F.~Liu, ``Oil price forecasting: A hybrid gru neural network based on decomposition--reconstruction methods,'' \emph{Expert Systems with Applications}, vol. 218, p. 119617, 2023.

\bibitem{radford2019language}
A.~Radford, J.~Wu, R.~Child, D.~Luan, D.~Amodei, I.~Sutskever \emph{et~al.}, ``Language models are unsupervised multitask learners,'' \emph{OpenAI blog}, vol.~1, no.~8, p.~9, 2019.

\bibitem{dosovitskiy2020image}
A.~Dosovitskiy, ``An image is worth 16x16 words: Transformers for image recognition at scale,'' \emph{arXiv preprint arXiv:2010.11929}, 2020.

\bibitem{chen2023chatgpt}
Z.~Chen, L.~N. Zheng, C.~Lu, J.~Yuan, and D.~Zhu, ``Chatgpt informed graph neural network for stock movement prediction,'' \emph{arXiv preprint arXiv:2306.03763}, 2023.

\bibitem{yu2023temporal}
X.~Yu, Z.~Chen, Y.~Ling, S.~Dong, Z.~Liu, and Y.~Lu, ``Temporal data meets llm--explainable financial time series forecasting,'' \emph{arXiv preprint arXiv:2306.11025}, 2023.

\bibitem{financial_aid}
\BIBentryALTinterwordspacing
R.~K., B.~D., J.~Ortagus, K.~R., E.~A., L.~M., and C.~J., ``State financial aid dataset,'' 2023. [Online]. Available: \url{https://informedstates.org/state-financial-aid-dataset-download}
\BIBentrySTDinterwordspacing

\bibitem{gruver2024large}
N.~Gruver, M.~Finzi, S.~Qiu, and A.~G. Wilson, ``Large language models are zero-shot time series forecasters,'' \emph{Advances in Neural Information Processing Systems}, vol.~36, 2024.

\bibitem{tan2024language}
M.~Tan, M.~A. Merrill, V.~Gupta, T.~Althoff, and T.~Hartvigsen, ``Are language models actually useful for time series forecasting?'' \emph{arXiv preprint arXiv:2406.16964}, 2024.

\bibitem{jin2023time}
M.~Jin, S.~Wang, L.~Ma, Z.~Chu, J.~Y. Zhang, X.~Shi, P.-Y. Chen, Y.~Liang, Y.-F. Li, S.~Pan \emph{et~al.}, ``Time-llm: Time series forecasting by reprogramming large language models,'' \emph{arXiv preprint arXiv:2310.01728}, 2023.

\bibitem{liu2024taming}
P.~Liu, H.~Guo, T.~Dai, N.~Li, J.~Bao, X.~Ren, Y.~Jiang, and S.-T. Xia, ``Taming pre-trained llms for generalised time series forecasting via cross-modal knowledge distillation,'' \emph{arXiv preprint arXiv:2403.07300}, 2024.

\bibitem{zhou2023one}
T.~Zhou, P.~Niu, L.~Sun, R.~Jin \emph{et~al.}, ``One fits all: Power general time series analysis by pretrained lm,'' \emph{Advances in neural information processing systems}, vol.~36, pp. 43\,322--43\,355, 2023.

\bibitem{Zeng2022AreTE}
A.~Zeng, M.~Chen, L.~Zhang, and Q.~Xu, ``Are transformers effective for time series forecasting?'' in \emph{Proceedings of the AAAI Conference on Artificial Intelligence}, 2023.

\bibitem{liu2024itransformer}
\BIBentryALTinterwordspacing
Y.~Liu, T.~Hu, H.~Zhang, H.~Wu, S.~Wang, L.~Ma, and M.~Long, ``itransformer: Inverted transformers are effective for time series forecasting,'' in \emph{The Twelfth International Conference on Learning Representations}, 2024. [Online]. Available: \url{https://openreview.net/forum?id=JePfAI8fah}
\BIBentrySTDinterwordspacing

\bibitem{nie2022time}
Y.~Nie, N.~H. Nguyen, P.~Sinthong, and J.~Kalagnanam, ``A time series is worth 64 words: Long-term forecasting with transformers,'' \emph{arXiv preprint arXiv:2211.14730}, 2022.

\bibitem{wang2024timemixer}
S.~Wang, H.~Wu, X.~Shi, T.~Hu, H.~Luo, L.~Ma, J.~Y. Zhang, and J.~Zhou, ``Timemixer: Decomposable multiscale mixing for time series forecasting,'' \emph{arXiv preprint arXiv:2405.14616}, 2024.

\bibitem{jin2024time}
M.~Jin, H.~Tang, C.~Zhang, Q.~Yu, C.~Liu, S.~Zhu, Y.~Zhang, and M.~Du, ``Time series forecasting with llms: Understanding and enhancing model capabilities,'' \emph{arXiv preprint arXiv:2402.10835}, 2024.

\bibitem{liang2024foundation}
Y.~Liang, H.~Wen, Y.~Nie, Y.~Jiang, M.~Jin, D.~Song, S.~Pan, and Q.~Wen, ``Foundation models for time series analysis: A tutorial and survey,'' in \emph{Proceedings of the 30th ACM SIGKDD Conference on Knowledge Discovery and Data Mining}, 2024, pp. 6555--6565.

\bibitem{das2023decoder}
A.~Das, W.~Kong, R.~Sen, and Y.~Zhou, ``A decoder-only foundation model for time-series forecasting,'' \emph{arXiv preprint arXiv:2310.10688}, 2023.

\bibitem{ansari2024chronos}
A.~F. Ansari, L.~Stella, C.~Turkmen, X.~Zhang, P.~Mercado, H.~Shen, O.~Shchur, S.~S. Rangapuram, S.~P. Arango, S.~Kapoor \emph{et~al.}, ``Chronos: Learning the language of time series,'' \emph{arXiv preprint arXiv:2403.07815}, 2024.

\end{thebibliography}
